\documentclass[10pt,twocolumn,letterpaper]{article}

\usepackage{wacv}
\usepackage{times}
\usepackage{epsfig}
\usepackage{graphicx}
\usepackage{amsmath}
\usepackage{amssymb}
\usepackage{multirow}

\usepackage{diagbox}
\usepackage{makecell}
\usepackage{fixmath}
\usepackage{enumitem}
\usepackage{pifont}
\usepackage{tabularx}
\usepackage{footnote}
\usepackage[bookmarks=false]{hyperref}
\usepackage{sidecap}

\usepackage{amsmath,amsfonts,bm}

\def\eqref#1{equation~\ref{#1}}

\def\1{\bm{1}}

\DeclareMathAlphabet{\mathsfit}{\encodingdefault}{\sfdefault}{m}{sl}
\SetMathAlphabet{\mathsfit}{bold}{\encodingdefault}{\sfdefault}{bx}{n}

\DeclareMathOperator*{\argmin}{arg\,min}

\graphicspath{{./imgs/}} 
\newcommand{\xmark}{\ding{55}}%

\newcommand{\minus}{\scalebox{0.75}[1.0]{$-$}}

\wacvfinalcopy 

\def\wacvPaperID{921} 

\ifwacvfinal\fi
\begin{document}

\title{Multi-way Encoding for Robustness}

\author{Donghyun Kim \\
Boston University\\
{\tt\small donhk@bu.edu}
\and
Sarah Adel Bargal \\
Boston University \\
{\tt\small sbargal@bu.edu}
\and
Jianming Zhang \\
Adobe Research \\
{\tt\small jianmzha@adobe.com}
\and
Stan Sclaroff \\
Boston University \\
{\tt\small sclaroff@bu.edu}
\and
}

\maketitle
\ifwacvfinal\thispagestyle{empty}\fi

\begin{abstract}
   Deep models are state-of-the-art for many computer vision tasks including image classification and object detection. However, it has been shown that deep models are vulnerable to adversarial examples. We highlight how one-hot encoding directly contributes to this vulnerability and propose breaking away from this widely-used, but highly-vulnerable mapping. We demonstrate that by leveraging a different output encoding, multi-way encoding, we decorrelate source and target models, making target models more secure. Our approach makes it more difficult for adversaries to find useful gradients for generating adversarial attacks. We present robustness for black-box and white-box attacks on four benchmark datasets: MNIST, CIFAR-10, CIFAR-100, and SVHN. The strength of our approach is also presented in the form of an attack for model watermarking, raising challenges in detecting stolen models. 
\end{abstract}

\section{Introduction}
	Deep learning models are vulnerable to adversarial examples \cite{szegedy2013intriguing}. Evidence shows that adversarial examples are transferable \cite{papernot2016transferability,liu2016delving}. This weakness can be exploited even if the adversary does not know the target model under attack, posing severe concerns about the security of the models. This is because an adversary can use a substitute model for generating adversarial examples for the target model, also known as \textit{black-box} attacks.

	Black-box attacks such as gradient-based attacks \cite{goodfellow6572explaining, madry2017towards} rely on perturbing the input by adding an amount dependent upon the gradient of the loss function with respect to the input (input gradient) of a substitute model. An example adversarial attack is $x^{adv} = x + \epsilon sign(\nabla_x Loss(f(x))$, where $f(x)$ is the model used to generate the attack. This added ``noise'' can fool a model although it may not be visually evident to a human. The assumption of such gradient-based approaches is that the gradients with respect to the input, of the substitute and target models, are correlated. 
	
	Our key observation is that the setup of conventional deep classification frameworks aids in the correlation of such gradients, and thereby makes these models more susceptible to black-box-attacks. Typically, a cross-entropy loss, softmax layer, and one-hot vector  encoding for the target label are used when training deep models. These conventions constrain the encoding length and number of possible non-zero gradient directions at the encoding layer. This makes it easier for an adversary to pick a harmful gradient direction and perform an attack from a substitute model.

\begin{figure*}
		\centering
		\includegraphics[width=0.75\linewidth]{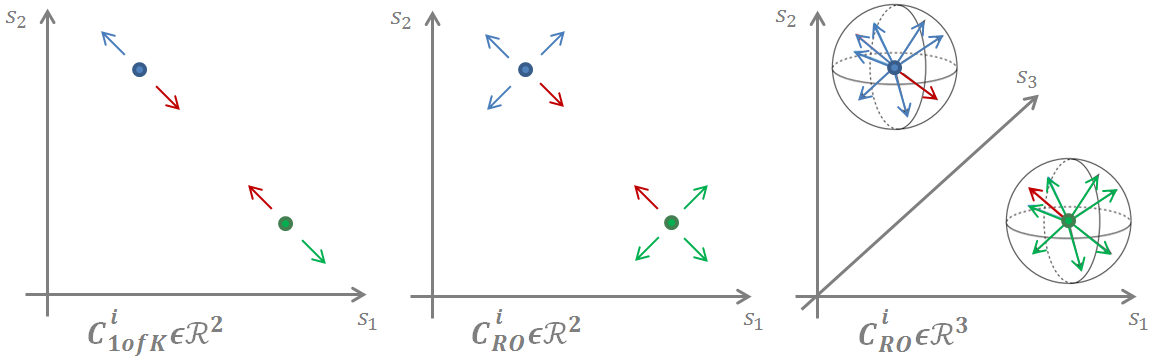} \\
		\hspace{3em} (a) \hspace{11em} (b) \hspace{11em} (c)
		\vspace*{+0.5em}
		\caption{Demonstration of the benefit of relaxing and increasing the encoding dimensionality, for a binary classification problem at the final encoding layer. $C_i$ is the codebook encoding for class $i$, axis $s_i$ represents the output activation of neuron $i$ in the output encoding layer, where $i = {1,\dots,l}$ and $l$ is the encoding dimensionality. The depicted points are correctly classified points of the green and blue classes. The arrows depict the possible non-zero perturbation directions $sign(\frac{\partial Loss}{\partial s_i})$. \textit{(a) $2D$ $1ofK$ softmax-crossentropy setup:} Only two non-zero gradient directions exist for a $1ofK$ encoding. Of these two directions, only one is an adversarial direction, depicted in red. \textit{(b) $2D$ multi-way encoding:} Four non-zero perturbation directions exist. The fraction of directions that now move a point to the adversarial class (red) drops. \textit{(c) $3D$ multi-way encoding:} A higher dimensional encoding results in a significantly lower fraction of gradient perturbations whose direction would move an input from the green ground-truth class to the blue class, or vice versa.
		}
		\label{fig:intro}
	\end{figure*}

	We aim to increase the adversarial robustness of deep models through \textit{model decorrelation}. Our multi-way encoding representation relaxes the one-hot encoding to a real number encoding, and embeds the encoding in a space that has a dimension that is higher than the number of classes. These encoding methods lead to an increased number of possible gradient directions, as illustrated in Fig.~\ref{fig:intro}. This makes it more difficult for an adversary to pick a harmful direction that would cause a misclassification. Multi-way encoding also helps improve a model's robustness in cases where the adversary has full knowledge of the target model under attack: a \textit{white-box} attack. 
	The benefits of multi-way encoding are demonstrated in experiments with four benchmark datasets.
	
	We also demonstrate the strength of \textit{model decorrelation} by introducing an attack for the recent model watermarking algorithm of Zhang \etal~\cite{zhang2018protecting}, which deliberately trains a model to misclassify certain watermarked images. We interpret such watermarked images as transferable adversarial examples. We demonstrate that the multi-way encoding reduces the transferability of the watermarked images. Our code is publicly available\footnote{\url{http://cs-people.bu.edu/donhk/research/Multiway\_encoding.html}}.
	
	We summarize our contributions as follows:
	\vspace{-0.55em}
	\begin{itemize}
		\item
		We propose a novel solution using multi-way encoding to alleviate the vulnerability caused by the $1ofK$ mapping through \textit{model decorrealtion}.
		\vspace{-0.2em}
		\item
		We empirically show that the proposed approach improves model robustness against both black-box attacks, white-box attacks, and general corruptions. 
		\vspace{-0.2em}
		\item 
		We also show the strength of our encoding by attacking a recently proposed model watermarking algorithm. 
	\end{itemize}

\section{Related Work}
	\textbf{Attacks.} Adversarial examples are crafted images for fooling a classifier with small perturbations. Recently, many different types of attacks have been proposed to craft adversarial examples. We focus on gradient-based attacks which deploy the gradient of the loss with respect to the input \cite{goodfellow6572explaining,kurakin2016adversarial,athalye2018obfuscated}. Goodfellow \etal~\cite{goodfellow6572explaining} propose the Fast Gradient Sign Method (FGSM) which generates adversarial images by adding the sign of the input gradients scaled by $\epsilon$, where the $\epsilon$ restricts $\ell_{\infty}$ of the perturbation. Kuerakin \etal~\cite{kurakin2016adversarial} propose the Basic Iterative Method (BIM), which is an iterative version of FGSM and is also called Projected Gradient Descent (PGD). Madry \etal~\cite{madry2017towards} show that PGD with randomly chosen starting points within allowed perturbation can make an attack stronger. Gradient-free attacks~\cite{brendel2017decision, li2019nattack, uesato2018adversarial} which do not use gradients from the target model can be used to check whether a defense relies on obfuscated gradients~\cite{athalye2018obfuscated}.
	
	\textbf{Defenses.} The goal of the defense is to make a correct prediction on adversarial examples. However, adversarial defenses can cause obfuscated gradients (\eg~\cite{xie2017mitigating, li2019nattack}) which are easily broken by Backward Pass Differentiable Approximation attack~\cite{athalye2018obfuscated}. Athalye \etal~\cite{athalye2018obfuscated} recommend performing several sanity tests to check obfuscated gradients for a defense. Madry \etal~\cite{madry2017towards} propose a defense based on the minimax formulation of adversarial training which has been extensively evaluated and justified. We also combine our method with the adversarial training and empirically show that our method does not rely on these fragile obfuscated gradients by following evaluations in \cite{athalye2018obfuscated}. However, the previous approach uses the conventional one-hot ($1ofK$) encoding for both source and target models, while we propose a higher dimensional multi-way encoding that obstructs the adversarial gradient search. Our goal is to mitigate the weakness of the transferability of adversarial examples by model decorrelation with our proposed encoding while not relying on obfuscated gradients and compromising white-box robustness at the same time. 
	
	\textbf{Output encoding.} There have been attempts to use alternate output encodings for image classification in deep models. Yang \etal~\cite{yang2015deep} and Rodriguez \etal~\cite{rodriguez2018beyond} use an output encoding that is based on Error-Correcting Output Codes (ECOC), for increased performance and faster convergence. In contrast, we use an alternate output encoding scheme, multi-way encoding, to make models more robust to adversarial attacks.
\section{Our Approach}
	
	In this section we will explain our approach using the following notation: $g(x)$ is the target model to be attacked, and $f(x)$ is the substitute model used to generate a black-box attack for $g(x)$. In the case of a white-box attack, $f(x)$ is $g(x)$. Canonical attacks like FGSM and PGD are gradient-based methods. Such approaches perturb an input $x$ by an amount dependent upon $sign(\nabla_x Loss(f(x)))$. An adversarial example $x^{adv}$ is generated as follows: 
	\begin{equation} \label{eq:1}
	x^{adv} = x + \epsilon sign(\nabla_x Loss(f(x))),
	\end{equation}
	where $\epsilon$ is the strength of the attack. Therefore $x^{adv}$ would be a translated version of $x$, in a vicinity further away from that of the ground-truth class, and thus becomes more likely to be misclassified, resulting in a successful adversarial attack. If the attack is a targeted one, $x$ could be deliberately moved towards some other specific target class. This is conventionally accomplished by using the adversarial class as the ground truth when back-propagating the loss, and subtracting the perturbation from the original input. The assumption being made in such approaches is that their input gradient direction is similar: $\nabla_x Loss(f(x)) \approx \nabla_x Loss(g(x)).$

	We now present the most widely used setup for training state-of-the-art deep classification networks comprising of one-hot encoding and softmax. Let the output activation of neuron $i$ in the final encoding (fully-connected) layer be $s_i$, where $i = {1, 2, \dots, k}$ and $k$ is the encoding length and the number of classes at the same time. Then, the softmax probability $y_i$ of $s_i$, and the cross-entropy loss are:
	\begin{equation}
	y_i = \frac{e^{s_i}}{\sum_{c=1}^{k}e^{s_c}}, \quad \text{and} \quad Loss = - \sum_{i=1}^{k} t_i log(y_i),
	\end{equation}
	respectively, where $t_i \in \{0, 1\}$ is the corresponding ground-truth one-hot vector encoding. The partial derivative of the loss with respect to the pre-softmax logit output is:
	\begin{equation} \label{eq:4}
	\frac{\partial Loss}{\partial s_i} = y_i - t_i.
	\end{equation}

	Combined with the most widely used one-hot ($1ofK$) encoding scheme, the derivative in Eq.~\ref{eq:4} makes the gradients of substitute and target models strongly correlated. We demonstrate this as follows:  
	Given a ground-truth example belonging to class [1,0,\dots,0], non-zero gradients of neuron 1 of the encoding layer will always be negative, while all other neurons will always be positive since $0 < y_i < 1$. So, regardless of the model architecture and the output, the signs of the partial derivatives are determined by the category, and thus the gradients for that category only lie in a limited hyperoctant (see Fig.~\ref{fig:intro} for the 2D case).
	{This constraint causes strong correlation in gradients in the final layer for different models using the $1ofK$ encoding. Our experiments suggest that this correlation can be carried all the way back to the input perturbations, making these models more vulnerable to attacks. }
	
	In this work, we aim to make $\nabla_x Loss(f(x))$ and $\nabla_x Loss(g(x))$ less correlated by encouraging \textit{model decorrelation}. We do this by introducing multi-way encoding instead of the conventional $1ofK$ encoding used by deep models for classification. Multi-way encoding significantly reduces the correlation between the gradients of the substitute and target models, making it more challenging for an adversary to create an attack that is able to fool the classification model.
	
	The multi-way encoding we propose in this work is the Random Orthogonal ($RO$) output vector encoding generated via Gram-Schmidt orthogonalization. Starting with a random matrix $\textbf{M}=[a_1|a_2|\dots|a_n] \in \mathbb{R}^{k \times l}$, the first, second, and $k^{th}$ orthogonal vectors are computed as follows: 
		\begin{align}
		\begin{split}
			u_1 &= a_1, \quad e_1 = \frac{u_1}{||u_1||}, \\
			u_2 &= a_2 \minus (a_2 \cdot e_1) e_1, \quad e_2 = \frac{u_2}{||u_2||}, \\
			u_k &= a_k \minus \dots \minus (a_k \cdot e_{k-1}) e_{k-1}, \quad e_k = \frac{u_k}{||u_k||}.
		\end{split}
	\end{align}
	For a classification problem of $k$ classes, we create a codebook $C_{RO} \in \mathbb{R}^{k \times l}$, where $C^i = \beta e_i$ is a length $l$ encoding for class $i$, and $i \in {1, \dots, k}$, and $\beta$ is a scaling hyper-parameter dependent upon $l$. A study on the selection of the length $l$ is presented in the experiments section.
	
	By breaking away from the $1ofK$ encoding, softmax and cross-entropy become ill-suited for the model architecture and training. Instead, we use the loss between the  output of the encoding-layer and the $RO$ ground-truth vector, $Loss(f(x),t_{RO})$, where $f(x) \in \mathbb{R}^{l}$ and $Loss$ measures the distance between $f(x)$ and $t_{RO}$. In our multi-way encoding setup, the final encoding ($s$) and $f(x)$ become equivalent. Classification is performed using $\arg\min_{i} Loss(f(x), t^i_{RO})$. We use Mean Squared Error (MSE) Loss.
	
	\renewcommand{\arraystretch}{1.1}
	\begin{table*}[t]
		\centering
		\begin{tabular}{l|c|c|c|c|c|c|c|c|c} 
			\hline
			 & 10  & 20 & 40 & 80 & 200 & 500 & 1000 & 2000 & 3000  \\
			\hline 
			Black-box & 45.4 &52.4 & 62.4 & 71.3 & 73.7 & 78.0 & 79.6 & 83.6 & 82.9\\
			\hline 
			White-box & 18.1 & 23.5 & 27.4 & 38.8 & 40.3 & 39.5 & 45.8 & 54.9 & 45.3 \\
			\hline
			Clean  & 96.8 &  97.0 & 97.9 & 98.3 &98.5 &98.8  &98.8 & 99.1 &98.9 \\
			\hline 
		\end{tabular}
		\vskip 0.1in
				\caption{This table presents the effect of increasing the dimension (10, 20, ..., 3000) of the output encoding layer of the multi-way encoding on the classification accuracy (\%) for MNIST on FGSM black-box, white-box attacks ($\epsilon=0.2$) and clean data. As the dimension increases, accuracy increases up to a certain point; We use 2000 for the length of our multi-way encoding layer.}
		\label{table:length_encoding}
	\end{table*}

	Fig.~\ref{fig:intro} illustrates how using the multi-way and longer encoding results in an increased number of possible gradient directions, reducing the probability of an adversary selecting a harmful direction that would cause misclassification. For simplicity we consider a binary classifier. Axis $s_i$ in each graph represents the output activation of neuron $i$ in the output encoding layer, where $i = {1,\dots,l}$. The depicted points are correctly classified points for the green and blue classes. The arrows depict the sign of non-zero gradients $\frac{\partial Loss}{\partial s_i}$. (a) Using a $1ofK$ encoding and a softmax-cross entropy classifier, there are only two directions for a point to move, a direct consequence of $1ofK$ encoding together with Eq.~\ref{eq:4}. Of these two directions, only one is an adversarial direction, depicted in red. (b) Using 2-dimensional multi-way encoding, we get four possible non-zero gradient directions. The fraction of directions that now move a correctly classified point to the adversarial class is reduced. (c) Using a higher dimension multi-way encoding results in a less constrained gradient space compared to that of $1ofK$ encoding. In the case of attacks formulated following Eq.~\ref{eq:1}, this results in $2^l$ possible gradient directions, rather than $l$ in the case of $1ofK$ encoding. The fraction of gradients whose direction would move the input from the green ground-truth class to the blue class, or vice versa, decreases significantly. In addition, multi-way encoding provides additional robustness by increasing the gradients' dimensionality. The effect of increasing dimensionality is shown in Table~\ref{table:length_encoding}.
	
	We also combine multi-way encoding with adversarial training for added robustness. We use the following formulation to solve the canonical min-max problem \cite{madry2017towards, kannan2018adversarial} against adversarial perturbations $\delta$ from PGD attacks:
	\begin{align}
		\begin{split}
			\argmin_\theta &[\mathbb{E}_{(x,y) \in {p}_{train}}\max_{\delta} (\mathcal{L}(\theta,x +\delta,y))\\&+\lambda \mathbb{E}_{(x,y)\in {p}_{train}}(\mathcal{L}(\theta,x,y))]
		\end{split}
	\end{align}
	where ${p}_{train}$ is the training data distribution, $(x,y)$ are the training points, and $\lambda$ determines a weight of the loss on clean data together with the adversarial examples at train time. For generating white-box adversarial attacks to our method, we minimize a variant of Carlini-Wagner (CW) loss~\cite{carlini2017towards}:\\
	\begin{equation}
	\label{eq:CW}
 	\max \left( \min _ { i \neq t } Loss \left( x ^ { }, e_i \right)  - Loss \left( x ^ {},  e_t \right) , -\kappa \right)
	\end{equation}  
	where $e_t$ is the ground-truth vector, $\kappa$ is a confidence, and $Loss$ is MSE loss. 

\section{Experiments}
	We conduct experiments on four commonly-used benchmark datasets: MNIST~\cite{lecun1998gradient}, CIFAR-10~\cite{krizhevsky2009learning}, CIFAR-100~\cite{krizhevsky2009learning}, and SVHN~\cite{netzer2011reading}. \textbf{MNIST} is a dataset of handwritten digits. It has a training set of 60K examples and a test set of 10K examples. 
	\textbf{CIFAR-10}  is a canonical benchmark for image classification and retrieval, with 60K images from 10 classes. The training set consists of 50K images, and the test set consists of 10K images. \textbf{CIFAR-100}  is similar to CIFAR-10 in format, but has 100 classes containing 600 images each. Each class has 500 training images and 100 testing images. \textbf{SVHN} is an image dataset for recognizing street view house numbers obtained from Google Street View images. The training set consists of 73K images, and the test set consists of 26K images.
	
	In this work we define a \textbf{\textit{black-box}} attack as one where the adversary knows the architecture and the output encoding used but not learned weights. We use two substitute models using $1ofK$ and $RO$ encodings respectively to evaluate our method. We define a \textbf{\textit{white-box}} attack as one where the adversary knows full information about our model, including the learned weights. The threat model is a $\ell_\infty$ bounded attack within the allowed perturbation $\epsilon$: 0.3 MNIST, 8/255.0 CIFAR-10, 8/255.0 CIFAR-100, 10/255.0 SVHN by following \cite{madry2017towards, buckman2018thermometer}.

	\begin{table}[t]
	\begin{center}
		\begin{tabular}{l|c|c|c}
			\hline
			\multirow{2}{*}{\textbf{Layer}} & \multicolumn{3}{c}{\textbf{Pearson Correlation Coefficient}} \\
			\cline{2-4}
			&  $A^{}_{1ofK}$, $A^{\prime}_{1ofK}$ & $A^{}_{RO}$, $A^{\prime}_{RO}$ & $A^{}_{RO}$, $A_{1ofK}$ \\
			\hline
			Conv1 & 0.29 & 0.06 & 0.0  \\
			\hline
			Conv2 & 0.24 & 0.15 & 0.01  \\ 
			\hline
			Input & 0.35 & 0.08 & 0.02\\  \hline
		\end{tabular}
		\vskip 0.1in
		
		\caption{Correlation of gradients between models of  different encodings. gradients of the loss with respect to the intermediate features of Conv1 and Conv2, and with respect to the input. Then, we compute the correlation coefficient of the sign of the gradients with respect to the intermediate features.}
		\label{tab:correlation}
	\end{center}
	
    \end{table}

	\subsection{Multi-way Encoding on MNIST}
	\label{sec1}
	In this section we provide an in-depth analysis of our multi-way encoding on the MNIST dataset. We conduct experiments to examine how multi-way output encodings can decorrelate gradients (Sec.~\ref{decorr}) and increase adversarial robustness (Sec.~\ref{rob}). We compare models trained on $1ofK$ output encodings with models having the same architecture but trained on multi-way output encodings. In all experiments we use $RO$ encoding as the multi-way encoding with dimension 2000 determined by Table \ref{table:length_encoding} and $\beta=1000$.  All models achieve ${\sim}99\%$ on the clean test set. Models A and C are LeNet-like CNNs and inherit their names from \cite{tramer2017ensemble}. We use their architecture with dropout before fully-connected layers.	We trained models A and C on MNIST with the momentum optimizer and an initial learning rate of 0.01, $momentum = 0.5$ with different weight initializations. It should be noted that, in this section, substitute and target models are trained on clean data and do not undergo any form of adversarial training.

	\begin{table*}[t]
	\centering
	\begin{tabular}{l|l|l|l|l|c} 
		\hline
		\diagbox{$g(x)$}{$f(x)$} & ${A}_{1ofK}$ & $A_{RO}$ & \textbf{$C_{1ofK}$} & \textbf{$C_{RO}$} & AVG BB\\
		\hline
		\textbf{$A_{1ofK}$} & 34.9 (1.00) * & 93.6 (0.02) & 56.8 (0.25) & 95.5 (0.03) & 82.0 \\
		\textbf{$A_{RO}$} & 88.7 (0.02) & 54.9 (1.00) * & 92.5 (0.02) & 82.9 (0.09) & 88.0 \\ 
		\textbf{$C_{1ofK}$} & 30.1 (0.25) & 83.6 (0.01) & 22.5 (1.00) * & 93.3 (0.01) & 69.0 \\ 
		\textbf{$C_{RO}$} & 94.3 (0.03) & 87.5 (0.09) & 96.1 (0.01) & 70.5 (1.00) * & 92.6\\ 
		\hline
	\end{tabular}
	\vskip 0.1in
		\caption{This table presents the classification accuracy (\%) of MNIST on FGSM black-box, white-box attacks, and average black-box (AVG BB) using architectures A and C.  $f(x)$ is a substitute model and $g(x)$ is a target model.
		We conclude: (a) $g(x)$ using $1ofK$ is more vulnerable to black-box attacks than $g(x)$ using $RO$. (b)  For white-box attacks, $RO$ encoding leads to better accuracy compared to $1ofK$. (c) In brackets is the correlation coefficient of the input gradients of $g(x)$ and $f(x)$. $RO$ results in a lower correlation compared to $1ofK$.}
	\label{table:fine_tuning}
    \end{table*}
    

	\subsubsection{Model Decorrelation}
	\label{decorr}

\begin{figure}[t]
		\centering
		\includegraphics[width=0.9\linewidth]{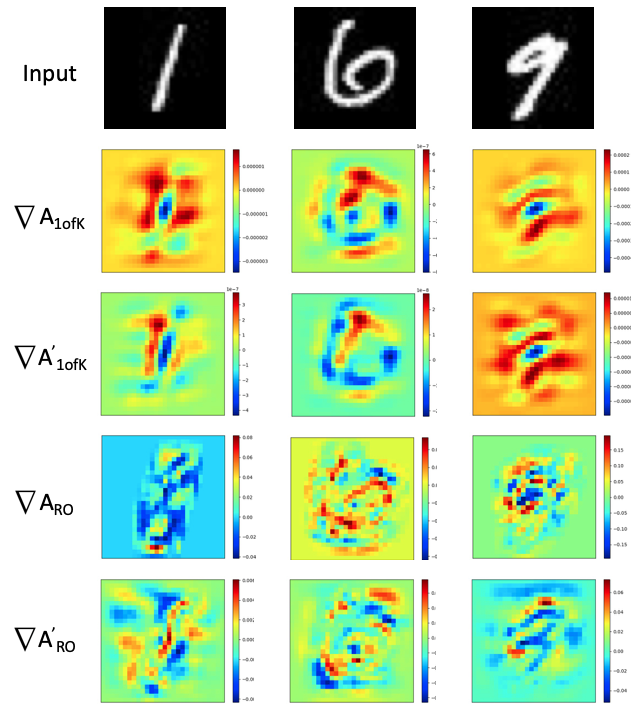} 
		\caption{Three sample examples of gradients of the loss with respect to an input from $A_{1ofK}$, $A'_{1ofK}$, $A_{RO}$, and $A'_{RO}$ models. All networks are independently trained with different weight initializations. The gradients of $A_{1ofK}$ and $A'_{1ofK}$ become similar after training while the gradients of $A_{RO}$ and $A'_{RO}$ are dissimlar. 
		}
		\label{fig:decorrelation_saliencyMap}
	\end{figure}

	In this section we present how multi-way encoding results in gradient decorrelation. Fig.~\ref{fig:decorrelation_saliencyMap} visualizes the value of the gradients of the loss with respect to input from the models: $A_{1ofK}$, $A'_{1ofK}$, $A_{RO}$, and $A'_{RO}$, for three sample images from the MNIST dataset. We observe that the gradients of $A_{1ofK}$ and $A'_{1ofK}$ are more similar than those of $A_{RO}$ and $A'_{RO}$. Also, the gradients of $1ofK$ encoding models are quite dissimilar compared to those of $RO$ encoding models.

	While Fig.~\ref{fig:decorrelation_saliencyMap} depicts three sample examples, we now present aggregate results on the entire MNIST dataset. We measure the correlation of gradients between all convolutional layers and the input layer of models trained on different encodings. We first compute the gradients of the loss with respect to intermediate features of Conv1 and Conv2. Then, we compute the Pearson correlation coefficient ($\rho$) of the sign of the gradients with respect to the intermediate features between models based on the following equation:
	For further comparison, we train models $A^{\prime}_{1ofK}$ and $A^{\prime}_{RO}$, which are independently initialized from $A_{1ofK}$ and $A_{RO}$. We average gradients of convolutional layers over channels in the same way a gradient-based saliency map is generated \cite{selvaraju2017grad}. Otherwise, the order of convolutional filters affects the correlations and makes it difficult to measure proper correlations between models. In this sense, the correlations at FC layers do not give meaningful information since neurons in the FC layer do not have a strict ordering.
	
	In Table~\ref{tab:correlation}, we find that the correlations of Conv1 and Conv2 between $1ofK$ models are much higher than those of $RO$ models. Table~\ref{tab:correlation} also shows that the correlations between $RO$ and $1ofK$ are also low. In addition, $RO$ models are not highly correlated even though they are using the same encoding scheme. At the input layer, the correlations between $RO$ and $1ofK$ are almost zero, but $1ofK$ models have a significantly higher correlation.
	
	We present ablation studies on our method in Section A in the supplementary material.
	
    \begin{table*}[t]
		\centering
		\begin{tabular}{ll|c|c|c|c|c|c} 
			\hline 
			{\multirow{2}{*}{\textbf{\textbf{Dataset}}}} & {\multirow{2}{*}{\textbf{\textbf{Model}}}} & \multicolumn{6}{c}{\textbf{Accuracy (\%)}}  \\ 
			\cline{3-8}
			&  & \makecell{Blackbox \\ \#steps:1K \\ ($1ofK$)} & \makecell{Blackbox  \\ \#steps:1K \\ ($RO$)} &  \makecell{Whitebox\\ \#steps:1K \\ } & \makecell{Whitebox\\\#steps:5K\\50-restarts} & \makecell{Gradient -free\\ ($\ell_2$, $\ell_\infty$~\cite{brendel2017decision,li2019nattack})}  & Clean\\ 
			\hline 
			\multirow{2}{*}{MNIST} & Madry \etal~ \cite{madry2017towards} & 94.9 & 95.0  & 92.2 & 89.6  & 35.4 &  98.4 \\
			& \textbf{Ours} & \textbf{96.9} & \textbf{96.8} &  \textbf{94.9} &\textbf{94.1} &  \textbf{42.7} & \textbf{99.2} \\
			\hline 
			\multirow{2}{*}{CIFAR-10} &  Madry \etal~ \cite{madry2017towards} & 62.5 &  \textbf{73.8} & 45.3  & 44.9 & 46.6 & 87.3 \\
			& \textbf{Ours} & \textbf{65.2} & 72.2 & \textbf{53.1} & \textbf{52.4} & \textbf{56.9} & \textbf{89.4} \\
			\hline 
		\end{tabular}
		\vskip 0.1in
		\caption{Comparison against the released models of Madry \etal~\cite{madry2017towards} on white-box, black-box PGD attacks, gradient-free attacks and on clean data. We report results on white and black-box PGD attacks generated using the 1K iterations. We observe that our approach is more resilient to these types of attacks and obtains improvements on clean data.}
		\label{table:adv_training}
	\end{table*}

	\begin{table*}[t]
		\begin{tabularx}{\textwidth}{c|cccccccccccc|}
			\hline
			\multirow{2}{*}{Model} & \multicolumn{11}{c}{Corruptions}       \\
			& AVG & Bright.     & Spatter & Jpeg  & Elestic & Motion        & Zoom     & Impulse  & Speckle & Gauss. noise& Snow  \\ \hline
			Madry \etal  &81.5     & 87.1           & 81.6    & 85.4  & 81.7    & 80.4          & 82.7     & 68.8     & 81.8 & 82.2   & 82.6     \\
			Ours             &\textbf{83.4}      & \textbf{89.3}           & \textbf{84.4}    & \textbf{87.1}  & \textbf{83.2}    & \textbf{81.6}          & \textbf{84.0}     & \textbf{72.0}     & \textbf{84.3}   & \textbf{84.3} & \textbf{84.7}\\
			\hline
			
		\end{tabularx}
				\vskip 0.1in
				\caption{Evaluation on common corruptions and perturbations on CIFAR-10~\cite{hendrycks2019benchmarking}. Our method obtains higher accuracy (test accuracy \%)  for all corruptions and perturbations. The first column (AVG) presents the averaged accuracy for all cases.}
		\label{table:corruption}
	\end{table*}
	\subsubsection{Robustness}
	\label{rob}
	
	Table \ref{table:fine_tuning} presents the classification accuracy (\%) of target models under attack from various substitute models. Columns represent the substitute models used to generate FGMSM attacks of strength $\epsilon = 0.2$ and rows represent the target models to be tested on the attacks. The diagonal represents white-box attacks  and others represent black-box attacks. Every cell corresponds to an attack from a substitute model $f(x)$ for a target model $g(x)$.  The last column reports the average accuracy on black-box attacks.
	
	By comparing the last column of Table \ref{table:fine_tuning}, the $g(x)$ using the $1ofK$ encoding is more vulnerable to black-box attacks than the corresponding $g(x)$ using the $RO$ encoding.  Black-box attacks become stronger if $f(x)$ uses the same encoding as $g(x)$. In addition, even though the same encoding is used, $RO$ models maintain higher robustness to the black-box attacks compared to $1ofK$ models (\eg~$82.9\%$ when $C_{RO}$ attacks$ A_{RO}$  vs. $56.8\%$ when $C_{1ofK}$ attacks $ A_{1ofK}$). This suggests that $RO$ encoding is more resilient to black-box attacks.
	
	It is also evident from the results of this experiment in Table \ref{table:fine_tuning} that even when the source and target models are the same, denoted by (*), $RO$ encoding leads to better accuracy, and therefore robustness to white-box attacks, compared to $1ofK$ encoding.	
	
	Finally, Table \ref{table:fine_tuning} reports the correlation coefficient of $sign(\nabla_x Loss(f(x)))$ and $sign(\nabla_x Loss(g(x)))$ in Eq. \ref{eq:1}. These gradients are significantly less correlated when the source and target models use different encodings. In addition, $RO$ results in a lower correlation compared to $1ofK$ when the same encoding is used in the source and target models.


	\subsection{Benchmark Results}
	\label{sec:2}
	
	In this section we analyze the case where we combine our method with adversarial training (Eq.~\ref{eq:CW}). We compare against the strong baseline of Madry \etal~\cite{madry2017towards}, which also uses adversarial training. For adversarial training, we use a mix of clean and adversarial examples for MNIST, CIFAR-10, and CIFAR-100, and adversarial examples only for SVHN following the experimental setup and the threat models used by Madry \etal~\cite{madry2017towards} and Buckman \etal~\cite{buckman2018thermometer}. We use PGD attacks with a random start, and follow the PGD parameter configuration of \cite{madry2017towards, kannan2018adversarial, buckman2018thermometer}. 
	
	We directly compare our method with the publicly released versions of Madry \etal~\cite{madry2017towards} on MNIST and CIFAR-10 in Table~\ref{table:adv_training}. We present results for 1K-step PGD black-box attacks generated from the independently trained copy of Madry \etal~(the first column) and the model trained with the RO (the second column). It should be noted that the substitute model uses the same RO encoding parameters used for the target model. We generate 1K-step PGD white-box attacks with a random start. In addition, we generate 5K-step PGD attacks with 50 random restarts.
	
	Table~\ref{table:adv_training} demonstrates the robustness of multi-way encoding for black-box attacks, while at the same time maintaining high accuracy for white-box attacks and clean data.
	
	The black-box attacks in the second column of Table~\ref{table:adv_training} show the robustness even when an adversary knows the exact value of the encoding used for the target model. We generate high-confidence PGD attacks with Eq.~\ref{eq:CW} from the independently trained copy of the $RO$ model. Our model achieves higher worst-case robustness compared to the baseline. In CIFAR-10, the black-box attacks from the $RO$ model (\ie~the substitute model uses the same $RO$ encoding parameters) are much weaker than the black-box attacks from the $1ofK$ model. This shows that $RO$ can effectively decorrelate the target model even when the encoding is exposed to an adversary. This is also consistent with the results where the correlation of gradients are lower in Table~\ref{tab:correlation} and the black-box robustness of the $RO$ models is higher in Table~\ref{table:fine_tuning} even when the substitute model uses the same $RO$ encoding as the target model.
	
	In the third and fourth columns, our defense also achieves higher robustness than the baseline on PGD white-box attacks. We observe that increasing random restarts decreases robustness on both ours and Madry \etal, which implies a gradient masking effect. However, this is due to the non-convexity of the loss landscape, so that this type of gradient masking can happen to all deep models. This type of gradient masking is different from obfuscated gradients~\cite{athalye2018obfuscated} which are easily broken by gradient-free attacks~\cite{li2019nattack}, which does not use the gradient from the target model. We show that this type of gradient masking is not easily broken by gradient-free  attacks in the fifth column of Table~\ref{table:adv_training}.
	
	\begin{figure*}[t]
		\centering
		
		\includegraphics[width=1.0\linewidth]{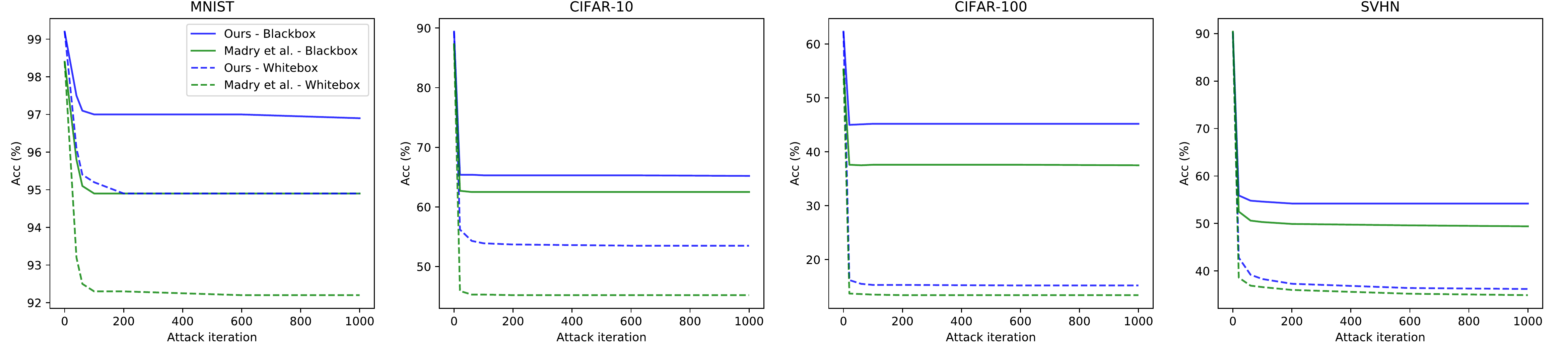} 
		\caption{ We generate PGD attacks with the different number of iterations and compare our method with the baseline by following~\cite{athalye2018obfuscated}. Black-box attacks are generated from an independently trained copy of the baseline. We observe that (1) the attack is converged; (2) iterative attacks are stronger than single-step attacks; (3) white-box attacks are stronger than black-box attacks.
		}
		\label{fig:iter}
	\end{figure*}
	
	We follow \cite{carlini2019evaluating, li2019nattack} and perform gradient-free attacks to check for signs of obfuscated gradients. In the fifth column, we evaluate our method on  gradient-free attacks; (1) decision-based attacks \cite{brendel2017decision, schott2018towards} for MNIST and (2) query-based attacks \cite{li2019nattack} using Eq.~\ref{eq:CW} for CIFAR-10. It should be noted the decision-based attacks are $\ell_{2}$ bounded which violates our threat model. 
	From the results, gradient-free attacks could not break our defense and we do not observe a sign of obfuscated gradients.
	
	\subsection{Further Analysis}
	 \textbf{Robustness on General Corruptions.} Recent works~\cite{ford2019adversarial, hendrycks2019benchmarking} highlight the close relationship between adversarial robustness and general corruption (\eg~Gaussian noise) robustness. It is observed that the certified defense~\cite{madry2017towards} increased robustness on not only adversarial examples but also corrupted examples, while failed defenses (\eg~\cite{xie2017mitigating, liao2018defense}) could not increase robustness on corrupted examples. Ford \etal~\cite{ford2019adversarial} argue that defense should have higher robustness on general corruptions and recommend reporting corruption robustness as a sanity check for a defense. We evaluate our method’s general robustness, using the dataset of Hendrycks \etal~\cite{hendrycks2019benchmarking} designed to test the common corruptions and perturbations on CIFAR-10. As reported in Table~\ref{table:corruption}, we achieve higher robustness in all cases.
	
	 \textbf{Additional Datasets.} We train the method of Madry \etal~and our approach on CIFAR-100 and SVHN. Fig.~\ref{fig:iter} represents the robustness on black-box and white-box attacks on different PGD attack iterations. We also report the clean accuracy when the iteration $=0$. Our method achieves higher robustness on the black-box and white-box attacks on CIFAR-100 and SVHN. In addition, we also improve the clean accuracy by $7\%$ on CIFAR-100. We observe that that attack success rates are converged with large iterations.
	
	 \textbf{Checking Obfuscated Gradients.} To check if our method relies on obfuscated gradients~\cite{athalye2018obfuscated}, we provide evaluations by following the guidelines of Athalye \etal~\cite{athalye2018obfuscated}. We include plots of the different PGD attack iterations in Fig.~\ref{fig:iter}. Transfer-based attacks can effectively check if a defense method relies on obfuscated gradients~\cite{athalye2018obfuscated}. Our model is robust to the transfer-based attacks than Madry \etal. We also observe that (1) the attack is converged; (2) iterative attacks are stronger than single-step attacks; (3) white-box attacks are stronger than black-box attacks. From these evaluations, we do not find a sign of obfuscated gradients and our method performs better than the baseline. Additional sanity check can be found in Section C of the supplementary.

	\textbf{Analysis on White-box Attacks.} 
\begin{figure}[t]
	\centering
	\includegraphics[width=1.0\linewidth]{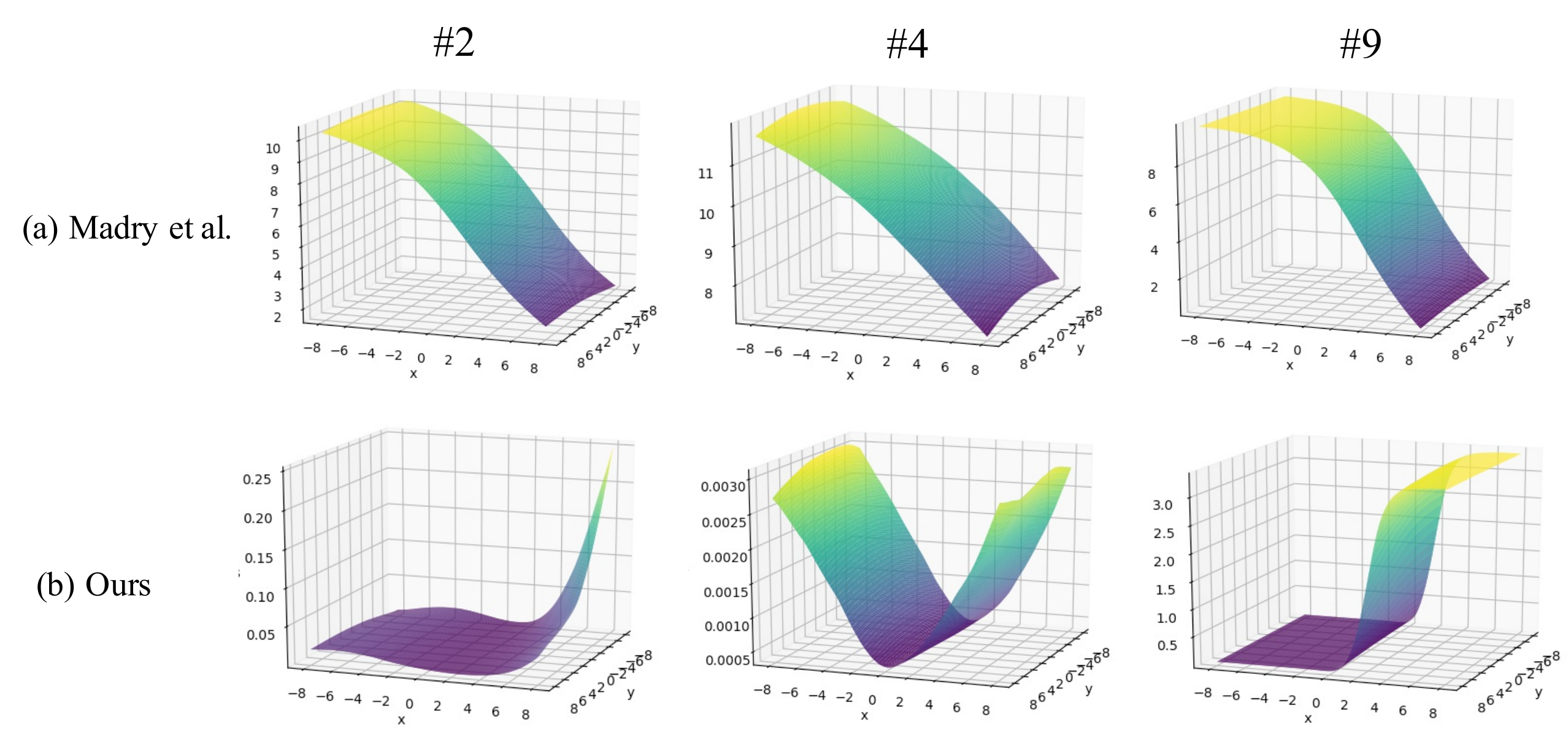} 
	\caption{Comparisons of landscapes of the baseline model of Madry \etal~\cite{madry2017towards} and our model. The x-axis represents the magnitude of the gradient direction of the loss w.r.t. input, the y-axis represents the magnitude of a random direction, and the z-axis represents (a) the value of the ground-truth neuron in the final layer for the baseline, and (b) the mean squared error loss between the final encoding and the ground-truth vector at each input data point $(x^{adv} = x' + x*r_1+ y *r_2)$. In this figure, we depict the landscapes of $x'$ of the three examples.}		
	\label{fig:landscape}
\end{figure}
The goal of this section is to analyze how our method improves white-box robustness. Following \cite{engstrom2018evaluating}, we plot the landscapes of the loss of our method and the value of neurons that correspond to the ground-truth class of the baseline model \cite{madry2017towards} for CIFAR-10. Since cross-entropy loss with softmax layer can be biased to the norm of neuron values at the final layer (Section 3 in \cite{carlini2016defensive}), we instead plot the value of the ground-truth neuron for the baseline model. In Fig.~\ref{fig:landscape}, The x-axis represents the magnitude of the direction of $r_1= sign(\nabla_{x'} Loss(f(x'))$ and the y-axis represents the magnitude of a random direction, $r _ { 2 } \sim \text { Rademacher } ( 0.5 )$. For our model, z-axis represents the mean squared error loss between the final encoding and the ground-truth vector at each input $ (x^{adv} = x' + x*r_1+ y *r_2)$. For Madry \etal, the z-axis represents the value of the ground-truth neuron in the final layer which is directly responsible for the loss. 

For Madry \etal, the landscapes show linearity along with the direction of $r_1$ for over the test set regardless of misclassification. However, our model shows non-linear behaviors over correctly classified examples but linear-like behaviors for misclassified examples. We show representative landscapes from the three test points. Fig.~\ref{fig:landscape} (a) shows the linearity along with the direction of $r_1$ for the baseline. The value of \#4 decreases slowly but still shows linear behavior and the values of \#2 and \#9 decrease significantly. \cite{goodfellow6572explaining} argue that the linearity is a primary cause of vulnerability. We also claim that the linear behavior makes it easier to find harmful gradients for the \textit{first-order adversary} like PGD attacks. Fig.~\ref{fig:landscape} (b), we observe that our model shows non-linear behavior with the direction of $r_1$ for the correctly classified examples: \#2, \#4. For \#9, the more linear-like behavior results in a much higher loss and misclassification. 


When ground-truth vectors are one-hot encodings, decreasing the output value of the neuron corresponding to the ground-truth class significantly would cause a misclassification. However, when ground-truth vectors are multi-way encodings, no single neuron is solely responsible for misclassification, but a more complex combination of neurons. Since the loss in our model is computed on multiple neurons at the final layer, an adversarial direction may increase the loss of certain neurons but it may also decrease the loss of other neurons at the same time. We argue that non-linearity is related to our high dimensional encoding layer which provides additional robustness to \textit{first-order white-box attacks} in addition to black-box attacks.
	
\section{Attacking Model Watermarking via Model Decorrelation}
Zhang \etal~\cite{zhang2018protecting} introduced an algorithm to detect whether a model is stolen or not. They do so by adding a watermark to sample images of specific classes and deliberately training the model to misclassify these examples to other specific classes. Even if their pre-trained model is stolen, the model should make a misclassification on the watermarked image. This approach has demonstrated to be robust even when the model is fine-tuned on a different training set.

We interpret the watermarked image used to deliberately cause a misclassification as a \textbf{\textit{transferable adversarial example}}. We introduce an attack for this algorithm using our multi-way encoding, making it more challenging to detect whether a model is stolen or not. We do this by fine-tuning the stolen model using multi-way encoding, rather than the encoding used in pre-training the model. We show that our multi-way encoding successfully decorrelates a model from the pre-trained model and, as a result, adversarial examples become less transferable.

We follow the same CIFAR-10 experimental setup for detecting a stolen model as in Zhang \etal: We split the test set into two halves. The first half is used to fine-tune pre-trained networks, and the second half is used to evaluate new models. 
When we fine-tune the $1ofK$ model, we re-initialize the last layer. When we fine-tune the $RO$ model we replace the output encoding layer with our 2000-dimension fully-connected layer, drop the softmax, and freeze convolutional weights.

We present results on the CIFAR-10 dataset in Table \ref{table:watermarking}. When the fine-tuning was performed using the $1ofK$ encoding (also used in pre-training the model), watermarking detection is 87.8\%, and when the fine-tuning was performed using the multi-way $RO$ encoding the watermarking detection is only 12.9\% while taking advantage of the pre-trained weights of the stolen model. The watermark detection rate of the model fine-tuned using $RO$ is significantly lower than that of model fine-tuned using $1ofK$ encoding, and is more comparable to models that are trained from scratch and do not use the stolen model (6.1\% and 10.0\%). These results suggest that our multi-way encoding successfully decorrelates the target model (finetuned $\textbf{Net}_{RO}$) from the source model ($\textbf{Stolen Net}_{1ofK}$).

\begin{table}
	\begin{center}
	
	\begin{tabular}{l|c|c|c} 
		\hline
		& \textbf{Finetune?} & \makecell{\textbf{Test Acc} \\ \textbf{(\%)}} & \makecell{\textbf{Watermark.} \\ \textbf{Acc} \textbf{(\%)}}\\ 
		\hline 
		$\textbf{StolenNet}_{1ofK} $ & \xmark & 84.7 & 98.6 \\
		\hline
		$\textbf{Net}_{1ofK}$ & \xmark & 48.3 & 6.1 \\
		$\textbf{Net}_{RO}$ & \xmark & 48.0 & 10.0 \\
		$\textbf{Net}_{1ofK}$ &  \checkmark & 85.6 & 87.8  \\
		$\textbf{Net}_{RO}$ & \checkmark & 80.2 & 12.9 \\
		\hline
    \end{tabular}
    \vskip 0.1in
    \caption{Our attack is capable of fooling the watermarking detection algorithm of~\cite{zhang2018protecting} via model decorrelation. Fine-tuning a stolen model using $RO$ encoding remarkably reduces the watermarking detection accuracy, and makes it comparable to the accuracy of models trained from scratch and do not use the stolen model. The accuracy of fine-tuned models benefits significantly from the pre-trained weights of the stolen model.}	
    \label{table:watermarking}
    \end{center}
\end{table}
\section{Conclusion}
	By relaxing the $1ofK$ encoding to a real number encoding, together with increasing the encoding dimensionality, our multi-way encoding decorrelates source and target models, confounding an attacker by making it more difficult to perturb an input in transferrable gradient direction(s) that would result in misclassification of a correctly classified example. We present stronger robustness on four benchmark datasets for both black-box and white-box attacks and we also improve classification accuracy on clean data. We demonstrate the strength of model decorrelation with our approach by introducing an attack for model watermarking, decorrelating a target model from the source model.

{\small
\bibliographystyle{ieee}
\bibliography{egbib}
}

\newpage

\appendix
\section*{Acknowledgments}
	We thank Kate Saenko, Vitaly Ablavsky, Adrian Vladu, Seong Joon Oh, Tae-Hyun Oh, and Bryan A. Plummer for helpful discussions. This work was supported in part by gifts from Adobe.
\section*{Appendix}

\begin{table*}[t!]
\begin{center}
\caption {This table presents black-box attacks from the substitute model $A_{1ofK}$ on various target models. $RO$ achieves the highest accuracy and the lowest input gradient correlation with the substitute model among the different target models.} 
\vskip 0.15in
\begin{tabular}{l|c|c|c|c|c|c}
\hline
\multicolumn{1}{c|}{\multirow{2}{*}{\begin{tabular}[c]{@{}c@{}}Target \\ Model\end{tabular}}} & \multicolumn{3}{c|}{A} & \multicolumn{3}{c}{C} \\ \cline{2-7} 
\multicolumn{1}{c|}{} & \multicolumn{1}{l|}{$RO_{softmax}$} & \multicolumn{1}{l|}{$1ofK_{MSE}$} & \multicolumn{1}{l|}{$RO$} & \multicolumn{1}{l|}{$RO_{softmax}$} & \multicolumn{1}{l|}{$1ofK_{MSE}$} & \multicolumn{1}{l}{$RO$} \\ \hline
Accuracy (\%) & 48.7 & 43.4 & 88.7 & 53.7 & 42.1 & 94.3 \\ \hline
\begin{tabular}[c]{@{}l@{}}Correlation\\ Coefficient\end{tabular} & 0.14 & 0.15 & 0.02 & 0.1 & 0.13 & 0.03 \\ \hline
\end{tabular}

\label{table:A1ofK_appendix}
\end{center}
\end{table*}

\begin{table*}[t!]
\begin{center}
\caption {This table presents black-box attacks from the substitute model $C_{1ofK}$ on various target models. $RO$ achieves the highest accuracy and the lowest input gradient correlation with the substitute model among the different target models.} 
\vskip 0.15in
\begin{tabular}{l|c|c|c|c|c|c}
\hline
\multicolumn{1}{c|}{\multirow{2}{*}{\begin{tabular}[c]{@{}c@{}}Target \\ Model\end{tabular}}} & \multicolumn{3}{c|}{A} & \multicolumn{3}{c}{C} \\ \cline{2-7} 
\multicolumn{1}{c|}{} & \multicolumn{1}{l|}{$RO_{softmax}$} & \multicolumn{1}{l|}{$1ofK_{MSE}$} & \multicolumn{1}{l|}{$RO$} & \multicolumn{1}{l|}{$RO_{softmax}$} & \multicolumn{1}{l|}{$1ofK_{MSE}$} & \multicolumn{1}{l}{$RO$} \\ \hline
Accuracy (\%) & 67.4 & 55.9 & 92.5 & 62.6 & 58.8 & 96.1 \\ \hline
\begin{tabular}[c]{@{}l@{}}Correlation\\ Coefficient\end{tabular} & 0.08 & 0.09 & 0.02 & 0.08 & 0.1 & 0.01 \\ \hline
\end{tabular}
\label{table:C1ofK_appendix}
\end{center}
\end{table*}
\section{Ablation Study on Encodings}

We perform ablation studies to further investigate the effectiveness of our $RO$ encoding. We train the model used in Table 2 in the original manuscript with two different combinations of encodings and loss functions. Please note that the two alternative models have 10 dimensions at the last layer while $RO$ has 2000 dimensions. 
\subsection{Alternative approach}
\subsubsection{$RO_{softmax}$} 
\label{A11}We evaluate a network that uses $RO$ encoding, a softmax layer, and cross-entropy loss. We compute the probability of $i^{th}$ class as follows:
$$ P(i|s) = \frac{\exp(\mathbf{s^\top e_i})}{\sum_{j=1}^{n} \exp(\mathbf{s^\top e_j})}$$
where $\mathbf{s}$ is the $\ell_2$ normalized final layer representation, $\mathbf{e_i}$ is the $RO$ encoding vector (ground-truth vector) from the codebook, and $n$ is the number of classes. 
\subsubsection{$1ofK_{MSE}$}
\label{A12}We also evaluate a network that uses mean-squared error (MSE) loss with the $1ofK$ encoding.
\subsection{Evaluation}
We generate FGSM attacks with $\epsilon=0.2$ from substitute models $A_{1ofK}$ and $C_{1ofK}$ on MNIST to evaluate the models of Section~\ref{A11} and Section~\ref{A12}. We also measure a correlation coefficient of the sign of the input gradients between target and substitute models as explained in Section $4.1.1$. Tables~\ref{table:A1ofK_appendix} and~\ref{table:C1ofK_appendix} demonstrate that $RO$, among the different target models, achieves the highest accuracy and the lowest input gradient correlation with the substitute model. It should be noted that the two alternative models have 10 neurons at the last layer while $RO$ has 2000 neurons. In addition, $RO_{softmax}$ has a softmax layer so that the gradients at the final layer are determined by a ground-truth class of an example.

\section{Transferability}
In Table 3 of the main paper, the black-box attacks of the second column report the robustness on black-box attacks from the independently trained copy of the $RO$ model. In this section, we analyze the black-box attack accuracy on CIFAR-10 by varying confidence $\kappa$ of Eq. 5 in the main paper. The higher confidence makes an attack to be more confident misclassification. We observe that the black-box attack accuracy converges at confidence$=1500$. We report the lowest accuracy in Table 3.

\begin{table*}[h]
    \centering
	\caption{This table presents accuracies on black-box attacks from $RO$ by varying confidence ($\kappa$). We generate 1000-step PGD attacks on CIFAR-10.} 
	\vskip 0.15in
	\begin{tabular}{l|l|l|l|l|l}
		\hline
		confidence       & 10    & 300    & 1500    & 3000   & 6000\\ \hline
		Accuracy (\%) & 83.0 & 80.9 & 72.2 & 72.2& 72.2 \\ \hline
	\end{tabular}
	\label{confidence}
\end{table*}

\section{Checking for Signs of Obfuscated Gradients}
In order to check if our method relies on obfuscated gradients~\cite{athalye2018obfuscated}, we report the accuracies on white-box attacks by varying epsilon on CIFAR-10 in Table \ref{bepsilon}. The maximum allowed perturbation for our model is 8/255, but we use larger epsilon to check the behavior of our model. We checked that increasing distortion bound monotonically increase attack success rates and unbounded attacks achieve $100\%$ attack success rate.

\begin{table*}[h]
\begin{center}
\caption {This table presents accuracies on white-box attacks by varying epsilon ($\ell_{\infty}$). Maximum allowed perturbation for our model is 8/255, but we use larger epsilon to check the behavior of our model.} 
\vskip 0.15in
\begin{tabular}{l|l|l|l|l|l|l|l|l|l}
\hline
epsilon       & 2    & 4    & 6    & 10   & 12   & 14   & 18   & 20 & Unbounded \\ \hline
Accuracy (\%) & 78.9 & 66.7 & 55.8 & 53.0 & 50.4 & 48.1 & 29.7 & 27 & 0         \\ \hline
\end{tabular}

\label{bepsilon}
\end{center}
\end{table*}

\end{document}